# Probabilistic Reasoning about Actions in Nonmonotonic Causal Theories

Thomas Eiter
Institut für Informationssysteme,
Technische Universität Wien
Favoritenstraße 9-11, A-1040 Vienna, Austria
eiter@kr.tuwien.ac.at

Thomas Lukasiewicz*
Dipartimento di Informatica e Sistemistica,
Università di Roma "La Sapienza"
Via Salaria 113, I-00198 Rome, Italy
lukasiewicz@dis.uniroma1.it

## Abstract

We present the language $P\mathcal{C}+$ for probabilistic reasoning about actions, which is a generalization of the action language $\mathcal{C}+$ that allows to deal with probabilistic as well as nondeterministic effects of actions. We define a formal semantics of $P\mathcal{C}+$ in terms of probabilistic transitions between sets of states. Using a concept of a history and its belief state, we then show how several important problems in reasoning about actions can be concisely formulated in our formalism.

## 1 INTRODUCTION

One of the most crucial problems that we have to face in reasoning about actions for mobile robotics in real-world environments is uncertainty, both about the initial situation of the robot's world and about the results of the actions taken by the robot (due to noisy effectors and/or sensors). One way of adding uncertainty to reasoning about actions is based on qualitative models in which all possible alternatives are equally taken into consideration. Another way is based on quantitative models where we have a probability distribution on the set of possible alternatives, and thus can numerically distinguish between possible alternatives.

Well-known first-order formalisms for reasoning about actions such as the situation calculus [18] easily allow for expressing qualitative uncertainty about the effects of actions and the initial situation of the world through disjunctive knowledge. Moreover, there are generalizations of the action language $\mathcal{A}$ [6] that allow for qualitative uncertainty in the form of nondeterministic actions. An important recent formalism in this family is the action language $\mathcal{C}+$ [7], which is based on the theory of nonmonotonic causal reasoning presented in [13], and has evolved from the action language $\mathcal{C}$ [8]. In addition to allowing for conditional and nondeterministic effects of actions, $\mathcal{C}+$ also supports concurrent actions as well as indirect effects and preconditions of actions through static causal laws. Closely related to it is the recent planning language $\mathcal{K}$ [3].

There are a number of formalisms for probabilistic reasoning about actions. In particular, Bacchus et al. [1] propose a probabilistic generalization of the situation calculus, which is based on first-order logics of probability, and which allows to reason about an agent's probabilistic degrees of belief and how these beliefs change when actions are executed. Poole's independent choice logic [16, 17] is based on acyclic logic programs under different "choices". Each choice along with the acyclic logic program produces a first-order model. By placing a probability distribution over the different choices, we then obtain a distribution over the set of first-order models. Other probabilistic extensions of the situation calculus are given in [12, 5]. A probabilistic extension of the action language $\mathcal{A}$ is given in [2].

The main idea behind the present paper is to orthogonally combine qualitative and quantitative uncertainty in a uniform framework for reasoning about actions: Even though there is extensive work on qualitative and quantitative models separately, there is only few work on such combinations. One such approach is due to Halpern and Tuttle [10], which combines nondeterminism and probabilistic uncertainty in a game-theoretic framework. Halpern and Tuttle argue in particular that "some choices in a distributed system must be viewed as inherently nondeterministic (or, perhaps better, nonprobabilistic), and that it is inappropriate, both philosophically and pragmatically, to model probabilistically what is inherently nondeterministic". This underlines the strong need for explicitly modeling qualitative uncertainty in addition to probabilistic uncertainty.

In this paper, we combine the qualitative uncertainty in the action language $\mathcal{C}+$ with probabilistic uncertainty as in [16, 17]. The main contributions are summarized as follows:

- We present the language $P\mathcal{C}+$ for probabilistic reasoning about actions, which is a probabilistic generalization of

---

*Alternate address: Institut für Informationssysteme, Technische Universität Wien, Favoritenstraße 9-11, A-1040 Vienna, Austria; e-mail: lukasiewicz@kr.tuwien.ac.at.

the action language $\mathcal{C}+$. It allows for representing actions with conditional and indirect effects, nondeterministic actions, and concurrently executed actions as the main features of $\mathcal{C}+$ as well as probabilistic knowledge about the effects of actions and the initial situation of the world.

- As a central property, $P\mathcal{C}+$ combines in a single framework qualitative as well as quantitative uncertainty, both about the effects of actions and about the initial situation of the world. Here, qualitative uncertainty is represented by forming a set of possible alternatives, while quantitative uncertainty is expressed through a probability distribution on a set of possible sets of possible alternatives.

- We define a formal semantics of $P\mathcal{C}+$ by interpreting probabilistic action descriptions in $P\mathcal{C}+$ as probabilistic transitions as in partially observable Markov decision processes (POMDPs) [11]. However, it is important to point out that these probabilistic transitions are between *sets of states* rather than *single states*. It is this which allows to handle qualitative uncertainty in addition to the quantitative probabilistic uncertainty as in POMDPs. Differently from standard POMDPs, our approach here only allows for observations without noise, but not for noisy sensing. It also does not deal with costs/rewards of actions.

- We define histories and their belief states in $P\mathcal{C}+$. Informally, a history $h$ is a sequence of actions and observations, which are labeled with a reasoning modality over sets of states. It has an associated belief state, which comprises possible sets of states and probabilistic information to describe the qualitative and quantitative knowledge about $h$. Note that such belief states model partial observability.

- We show how to express a number of important problems in probabilistic reasoning about actions (namely, the problems of prediction, postdiction, and planning; see especially [18, 12, 7]) in terms of belief states in $P\mathcal{C}+$.

The work closest in spirit to this paper is perhaps the one by Baral et al. [2], which also proposes a logic-based formalism for probabilistic reasoning about actions. However, there are several crucial differences. First, and as a central conceptual difference, our work orthogonally combines quantitative and qualitative uncertainty, in the form of probability distributions and sets of possible alternatives, respectively, while Baral et al. only allow for quantitative uncertainty in the form of probability distributions. Note that Baral et al. circumvent the problem of dealing with qualitative uncertainty by making the strong assumption of a uniform distribution whenever the probabilities for possible alternatives are not known. Second, Baral et al. allow only for a quite restricted form of probability distributions, which are either uniform distributions or produced from uniform distributions. Third, our language $P\mathcal{C}+$ generalizes the action language $\mathcal{C}+$, while Baral et al.'s language generalizes the action language $\mathcal{A}$. Note that $\mathcal{C}+$ is a novel action language that evolved from $\mathcal{A}$ and that is much more expressive than $\mathcal{A}$.

Another important formalism that is related to ours is Poole's independent choice logic [16, 17], which uses a similar way of adding probabilities to an approach based on acyclic logic programs. But also here the central conceptual difference is that Poole's independent choice logic does not allow for qualitative uncertainty in addition to quantitative uncertainty. Poole circumvents the problem of dealing with qualitative uncertainty by imposing the strong condition of acyclicity on logic programs. Furthermore, Poole's work is more inspired by the situation calculus and less by the action languages around $\mathcal{A}$.

The rest of this paper is organized as follows. In Section 2, we define $P\mathcal{C}+$ and its semantics in probabilistic transitions between sets of states. Section 3 introduces histories and belief states. In Section 4, we describe how several important problems in probabilistic reasoning about actions can be expressed in our framework. Section 5 summarizes the main results and gives an outlook on future research. Note that further technical results and detailed proofs of all results are given in the extended version of this paper [4].

## 2 SYNTAX AND SEMANTICS OF $P\mathcal{C}+$

In this section, we first recall nonmonotonic causal theories from [7]. We then present the language $P\mathcal{C}+$ for probabilistic reasoning about actions, and give an example of a probabilistic action description and initial database expressed in $P\mathcal{C}+$. We finally define the semantics of $P\mathcal{C}+$ through probabilistic transitions between sets of states.

Informally, the main idea behind our probabilistic extension of $\mathcal{C}+$ is to associate with the initial database and with every stochastic action a probability distribution on a set of *contexts*, which are values of exogenous variables. Every sequence of actions from an initial database is then associated with a probability distribution on a set of combined contexts. Hence, probabilistic reasoning about actions in $P\mathcal{C}+$ can essentially be reduced to standard reasoning about actions in $\mathcal{C}+$ with respect to such combined contexts. Note that Poole's independent choice logic [16, 17] uses a similar way of adding probabilities to an approach based on acyclic logic programs.

### 2.1 PRELIMINARIES

We now recall (nonmonotonic) causal theories from [7], which are used to specify initial sets of states and transitions from states to sets of states through actions. Roughly, a causal theory $T$ is a set of "causal rules" $F \Leftarrow G$ with the meaning "if $G$ holds, then there is a cause for $F$". In this paper, we consider only finite $T$. We now first define the syntax of causal theories and then their semantics.

We assume a finite set of *variables* $\mathcal{X}$. Every variable $X \in \mathcal{X}$ may take on *values* from a finite *domain* $I(X)$. We define *formulas* by induction as follows. *False* and *true*, denoted $\bot$ and $\top$, respectively, are formulas. If $X \in \mathcal{X}$ and $x \in I(X)$, then $X = x$ is a formula (called *atom*). If $F$ and $G$ are formulas, then also $\neg F$ and $(F \wedge G)$ are formulas. As usual, $X \neq x$ abbreviates $\neg X = x$. A *(causal) rule* is an expression of the form $F \Leftarrow G$, where $F$ and $G$ are formulas. A *causal theory* is a finite set of rules.

An *interpretation* $I$ of $\mathcal{Y} \subseteq \mathcal{X}$ maps every $Y \in \mathcal{Y}$ to an element of $I(Y)$. We use $I(\mathcal{Y})$ to denote the set of all interpretations of $\mathcal{Y}$. We obtain $I(F)$ and $I(F \Leftarrow G)$ from $F$ and $F \Leftarrow G$, respectively, by replacing every atom $Y = y$ such that $Y \in \mathcal{Y}$ and $I(Y) = y$ (resp., $I(Y) \neq y$) by $\top$ (resp., $\bot$). An interpretation $I$ of $\mathcal{Y}$ *satisfies* an atom $Y = y$ with $Y \in \mathcal{Y}$, denoted $I \models Y = y$, iff $I(Y) = y$. Satisfaction is extended to all formulas over $\mathcal{Y}$ as usual.

Let $T$ be a causal theory and $I$ be an interpretation of the variables in $T$. The *reduct* of $T$ relative to $I$, denoted $T^I$, is defined as $\{F \mid F \Leftarrow G \in T, I \models G\}$. The interpretation $I$ is a *model* of $T$, denoted $I \models T$, iff $I$ is the unique model of $T^I$. The theory $T$ is *consistent* iff it has a model.

## 2.2 SYNTAX OF P$\mathcal{C}$+

We next define the syntax of the probabilistic action language P$\mathcal{C}$+, which generalizes the syntax of $\mathcal{C}$+ [7]. We refer to [7] for further motivation and background for $\mathcal{C}$+. We illustrate the language P$\mathcal{C}$+ along a (simplistic) robot action domain, which is summarized in Fig. 1. This example shows in particular how quantitative as well as qualitative uncertainty about both the effects of actions and the initial situation of the world can be encoded in P$\mathcal{C}$+.

We divide the variables in $\mathcal{X}$ into *rigid*, *fluent*, *action*, and *context* variables. The fluent variables (or *fluents*) are additionally divided into *simple* and *statically determined* ones. We assume that action variables have the Boolean domain $\{\bot, \top\}$. Intuitively, the world is described through rigid variables and fluents. The values of rigid variables do not change when actions are performed, while those of simple (resp., statically determined) fluents may directly (resp., indirectly) change through actions. Action variables are used to describe actions, while context variables allow for adding probabilistic knowledge about the effects of actions and about the initial situation of the world.

**Example 2.1** In the robot action domain in Fig. 1, a mobile robot $r$ may move to the locations $a$, $b$, and $c$, and carry one of two objects $o_1$ and $o_2$ after pickup. This world is described through the simple fluents $at(o_1)$, $at(o_2)$, and $at(r)$ with the domain $\{a, b, c, lost\}$, where $at(O) = L$ iff $O$ is at location $L$. Moreover, we have the simple fluent $holds$ with the domain $\{o_1, o_2, nil\}$, where $holds = O$ iff $r$ holds $O$. We then have the action variables $goto(a)$, $goto(b)$, $goto(c)$, $pickup$, and $drop$, which represent the elementary actions "move to location $L$", "pick up an object", and "drop an object", respectively. Finally, the action "move to location $L$" succeeds only with a certain probability. To model this, we use the context variables $c_g(a)$, $c_g(b)$, and $c_g(c)$ with the domain $\{ok, fail\}$, where $c_g(L) = ok$ iff "move to location $L$" succeeds.

We next define static causal laws, which represent knowledge about fluents and rigid variables. Formally, a *static causal law* is an expression of the form

$$\textbf{caused } F \textbf{ if } G, \qquad (1)$$

where $F$ and $G$ are formulas such that either (a) every variable occurring in $F$ is a fluent, and no variable occurring in $G$ is an action variable, or (b) every variable occurring in $F$ or $G$ is rigid. If $G = \top$, then (1) is abbreviated by **caused** $F$. Roughly, (1) encodes that every state of the world that satisfies $G$ should also satisfy $F$. More formally, (1) is interpreted as the causal rule $F \Leftarrow G$.

**Example 2.2** The static causal law **caused** $at(O) = L$ **if** $holds = O \wedge at(r) = L$ (6) expresses that if the robot $r$ is at location $L$, and $r$ holds $O$, then $O$ is also at location $L$.

We now define dynamic causal laws, which express how the simple fluents change through actions, and which also encode execution denials for actions. Formally, a *dynamic causal law* is an expression of the form

$$\textbf{caused } F \textbf{ if } G \textbf{ after } H, \qquad (2)$$

where $F$, $G$, and $H$ are formulas such that (i) every variable occurring in $F$ is a simple fluent, (ii) no variable in $G$ is an action variable, and (iii) no variable in $H$ is a context variable. If $G = \top$, then (2) is abbreviated by **caused** $F$ **after** $H$. We use **inertial** $X$ to abbreviate the set of all rules (2) such that $F = G = H = X = x$ and $x \in I(X)$. If $F = \bot$ and $G = \top$, then (2) is called an *execution denial* and abbreviated by

$$\textbf{nonexecutable } H. \qquad (3)$$

Roughly, (2) expresses that every possible next state of the world that satisfies $G$ should also satisfy $F$, if the current state and the executed action satisfy $H$. More formally, (2) is interpreted as the causal rule $F \Leftarrow G \wedge H$, where $G$ and $F$ refer to the possible next states of the world, and $H$ refers to the current state and the executed action.

**Example 2.3** The dynamic causal law **caused** $holds = nil$ **after** $drop$ (13) says that $r$ holds nothing after $drop$. The execution denial (12) expresses that $pickup$ cannot be executed if $r$ already holds an object or if there is no object at the same location as $r$. The dynamic causal law (11) says that $r$ cannot pick up an object that is not at the same location as $r$, and (10) says that $r$ holds $o_1$ respectively $o_2$ after $pickup$. Thus, there is qualitative uncertainty in the effects of $pickup$: if both $o_1$ and $o_2$ are at the same location as $r$, then $pickup$ results in $r$ picking up either $o_1$ or $o_2$, but it is unpredictable which object $r$ actually picks up.

A *causal law* (or *axiom*) is a static or dynamic causal law. Our causal laws generalize their classical counterparts from [7] in the sense that they may also contain context variables. We next introduce the new concept of a context law. Context variables along with such context laws allow for expressing probabilistic effects of actions and probabilistic knowledge about the initial situation of the world.

More formally, a *dynamic context law* for a context variable $X \in \mathcal{X}$ is an expression of the form

$$X = (x_1 : p_1, \ldots, x_n : p_n) \textbf{ after } A, \qquad (4)$$

where (i) $I(X) = \{x_1, \ldots, x_n\}$, (ii) $p_1, \ldots, p_n > 0$, (iii) $p_1 + \cdots + p_n = 1$, and (iv) $A$ is a formula over action variables. We use $Pr(X = x_i)$ to denote $p_i$. If $A = \top$, then (4) is called a *static context law* and abbreviated by

$$X = (x_1 : p_1, \ldots, x_n : p_n). \qquad (5)$$

Roughly, (4) encodes that after executing an action that satisfies $A$, the probability that $X$ has the value $x_i$ is given by $p_i$. Note that a possible generalization of context laws could be to specify a set of probability distributions rather than a single probability distribution.

**Example 2.4** The actions $goto(a)$, $goto(b)$, and $goto(c)$ succeed only with certain probabilities. This is modeled using the context variables $c_g(a)$, $c_g(b)$, and $c_g(c)$ in the dynamic causal laws (7) and (8), along with the dynamic context laws (14)–(16). For example, the probability that $r$ really arrives at $a$ after executing $goto(a)$ is given by 0.95.

We next define the concept of a probabilistic action description (resp., initial database), which encodes the effects of all actions (resp., the initial situation of the world).

**Definition 2.5** A *probabilistic action description* $D$ is a finite set of causal and dynamic context laws such that $D$ contains exactly one dynamic context law for every context variable $X \in \mathcal{X}$ in $D$. Any such $D$ is a *probabilistic initial database*, if all causal and context laws in $D$ are static.

**Example 2.6** In the robot action domain in Fig. 1, the probabilistic action description $D$ is completely given by the sentences (6)–(17). Here, axioms (6) and (17) take care of the well-known ramification and frame problem, respectively. Note that axiom (18) forbids concurrent actions.

A probabilistic initial database $D_0$ may be given as follows. The initial locations of $o_1$ and $o_2$ are known with probabilities, which we express by the context variables $c_{at(o_1)}$ and $c_{at(o_2)}$, the static context laws $c_{at(o_1)} = (a:0.1, b:0.8, c:0.1)$ and $c_{at(o_2)} = (a:0.3, b:0.6, c:0.1)$, respectively, and the static causal laws **caused** $at(O) = L$ **if** $c_{at(O)} = L$, where $O \in \{o_1, o_2\}$ and $L \in \{a, b, c\}$. For example, object $o_1$ is at location $a$ with a probability of 0.1. Moreover, $r$ is at $a$ or $b$, expressed by **caused** $at(r) = a \vee at(r) = b$. Finally, $r$ holds no object, expressed by **caused** $holds = nil$.

---

(i) simple fluents: $at(O)$, $O \in \{o_1, o_2, r\}$: $\{a, b, c, lost\}$;
    $holds$: $\{o_1, o_2, nil\}$.

(ii) action variables: $goto(L)$, $L \in \{a, b, c\}$; $pickup$; $drop$.

(iii) context variables: $c_g(L)$, $L \in \{a, b, c\}$: $\{ok, fail\}$.

(iv) static causal laws:
$$\textbf{caused } at(O) = L \textbf{ if } holds = O \wedge at(r) = L \qquad (6)$$

(v) dynamic causal laws:
$$\textbf{caused } at(r) = L \textbf{ if } c_g(L) = ok \textbf{ after } goto(L) \qquad (7)$$
$$\textbf{caused } at(r) = L' \textbf{ if } at(r) = L' \wedge c_g(L) = fail \qquad (8)$$
$$\textbf{after } goto(L) \qquad \text{(for } L' \neq L\text{)}$$
$$\textbf{nonexecutable } goto(L) \wedge at(r) = L \qquad (9)$$

$$\textbf{caused } holds = O \textbf{ if } holds = O \qquad (10)$$
$$\textbf{after } pickup \qquad \text{(for } O \in \{o_1, o_2\}\text{)}$$
$$\textbf{caused } \bot \textbf{ if } holds = O \qquad (11)$$
$$\textbf{after } pickup \wedge at(r) = L \wedge at(O) \neq L$$
$$\textbf{nonexecutable } pickup \wedge [holds \neq nil \qquad (12)$$
$$\vee (at(r) = L \wedge at(o_1) \neq L \wedge at(o_2) \neq L)]$$
$$\textbf{caused } holds = nil \textbf{ after } drop \qquad (13)$$

(vi) dynamic context laws:
$$c_g(a) = (ok:0.95, fail:0.05) \textbf{ after } goto(a) \qquad (14)$$
$$c_g(b) = (ok:0.95, fail:0.05) \textbf{ after } goto(b) \qquad (15)$$
$$c_g(c) = (ok:0.90, fail:0.10) \textbf{ after } goto(c) \qquad (16)$$

(vii) inertial laws: for all simple fluents $f$:
$$\textbf{inertial } f \qquad (17)$$

(viii) other execution denials: for all action variables $a_1 \neq a_2$:
$$\textbf{nonexecutable } a_1 \wedge a_2 \qquad (18)$$

Figure 1: Robot Action Domain

In the sequel, $D$ (resp., $D_0$) denotes a probabilistic action description (resp., probabilistic initial database).

### 2.3 SEMANTICS OF P$\mathcal{C}$+

We now define the semantics of P$\mathcal{C}$+. Informally, certain interpretations of rigid and fluent variables serve as possible states of the world. We then associate with $D_0$ a collection of sets of such states, where each set of states has an associated probability. Furthermore, we associate with $D$ a mapping that assigns to each pair $(S, \sigma)$, consisting of a current set of states $S$ and a labeled action or observation $\sigma$, a probability distribution on a collection of future sets of states. Thus, we interpret $D$ by probabilistic transitions as in partially observable Markov decision processes (POMDPs) [11]. But the probabilistic transitions here are between *sets of states* rather than *single states*, which allows to handle qualitative uncertainty in addition to quantitative probabilistic uncertainty. Actions and observations are treated in a uniform way and labeled with modalities to specify how their preconditions (resp., observed formulas) are evaluated on sets of states (see also Section 4).

**Semantics of $D_0$.** Informally, we associate with $D_0$ a finite set of *contexts* $\gamma$, where every context $\gamma$ is in turn associated with a probability value $Pr_0(\gamma)$ and a set of states $\Phi_\gamma$. Thus, $D_0$ is interpreted as the collection of all $\Phi_\gamma$, where each $\Phi_\gamma$ has the probability $Pr_0(\gamma)$. We say that $D_0$ is *consistent* iff $\Phi_\gamma \neq \emptyset$ for all contexts $\gamma$.

Formally, let $\mathcal{X}_0$ denote the set of all context variables in $D_0$. We call $\gamma \in I(\mathcal{X}_0)$ a *context* for $D_0$. Its *probability*, denoted $Pr_0(\gamma)$, is defined as $\Pi_{X \in \mathcal{X}_0} Pr(X = \gamma(X))$. For $\mathcal{X}_0 = \emptyset$, the empty mapping $\gamma = \emptyset$ is the only context for $D_0$, which has the probability $Pr_0(\gamma) = 1$.

For each $\gamma \in I(\mathcal{X}_0)$, we define $\Phi_\gamma$ as the set of all models over rigid and fluent variables of the causal theory comprising all $\gamma(F \Leftarrow G)$ for each axiom (1) in $D_0$ and all $X = x \Leftarrow X = x$ for each simple fluent $X \in \mathcal{X}$ and $x \in I(X)$.

In order to define the semantics of $D$, we now formally define states, actions, and observations. We also define how a context $\gamma$, current state $s$, and action or observation $\sigma$ is associated with a set of future states $\Phi_\gamma(s, \sigma)$.

**States.** A *state* $s$ is either a member of some $\Phi_\gamma$, or a model over rigid and fluent variables of the causal theory comprising all $\gamma(F \Leftarrow G)$ for each axiom (1) in $D$, for any interpretation $\gamma$ of the context variables in $\mathcal{X}$, and all $X = x \Leftarrow X = x$ for each simple fluent $X \in \mathcal{X}$ and $x \in I(X)$.

**Actions.** An *action* $\alpha$ is an interpretation of the action variables in $\mathcal{X}$. Intuitively, each action variable is a basic action, and $\alpha$ is the concurrent execution of all basic actions that are true under $\alpha$. The *precondition* for $\alpha$, denoted $\pi_\alpha$, is the conjunction of all $\neg H$ for every execution denial (3) in $D$ such that $s \cup \alpha \models H$ for some state $s$. An action $\alpha$ is *executable* in a state $s$, denoted $\pi_\alpha(s)$, iff $s \cup \alpha \models \pi_\alpha$.

We next associate with every action $\alpha$ a set of contexts, and with each such context $\gamma$ a probability $Pr_\alpha(\gamma)$ and a mapping from states $s$ to a set of future states $\Phi_\gamma(s, \alpha)$. Intuitively, if $\alpha$ is executed in the state $s$ under the context $\gamma$, then the set of future states is given by $\Phi_\gamma(s, \alpha)$.

Formally, for states $s$ such that $\pi_\alpha(s)$, denote by $\mathcal{X}_{s,\alpha}$ the set of all context variables in some axiom (1) or (2) in $D$ such that $s \cup \alpha \models H$. We define $\mathcal{X}_\alpha$ as the union of all $\mathcal{X}_{s,\alpha}$. We call $\gamma \in I(\mathcal{X}_\alpha)$ a *context* for $\alpha$. Its *probability*, denoted $Pr_\alpha(\gamma)$, is defined as $\Pi_{X \in \mathcal{X}_\alpha} Pr(X = \gamma(X))$.

An *action transition* is a triple $(s, \alpha, s')$, where $s$ and $s'$ are states such that $s(X) = s'(X)$ for every rigid variable $X \in \mathcal{X}$, and $\alpha$ is an action such that $\pi_\alpha(s)$. A formula $F$ is *caused* in $(s, \alpha, s')$ under $\gamma \in I(\mathcal{X}_\alpha)$ iff $D$ contains either (a) an axiom (1) such that $s' \cup \gamma \models G$, or (b) an axiom (2) such that $s \cup \alpha \models H$ and $s' \cup \gamma \models G$. We say $(s, \alpha, s')$ is *causally explained* under $\gamma$ iff $s'$ is the only interpretation that satisfies all formulas caused in $(s, \alpha, s')$ under $\gamma$. For every state $s$ and action $\alpha$, we define $\Phi_\gamma(s, \alpha)$ as the set of all $s'$ such that $(s, \alpha, s')$ is *causally explained* under $\gamma$. Note that $\Phi_\gamma(s, \alpha) = \emptyset$ if no such $(s, \alpha, s')$ exists, in particular, if the action $\alpha$ is not executable in the state $s$.

**Observations.** An *observation* $\omega$ is a formula over fluents. For states $s$, we use $\pi_\omega(s)$ to abbreviate $s \models \omega$. In order to treat actions and observations in a uniform way, we also associate with every observation $\omega$ a set of contexts, and with each such context $\gamma$ a probability $Pr_\omega(\gamma)$ and a mapping from states $s$ to a set of future states $\Phi_\gamma(s, \omega)$.

Formally, we define $\mathcal{X}_\omega = \emptyset$. The empty mapping $\gamma \in I(\mathcal{X}_\omega)$ is the *context* for $\omega$. It has the *probability* $Pr_\omega(\gamma) = 1$.

For states $s$ and observations $\omega$, we define $\Phi_\gamma(s, \omega) = \{s\}$, if $\pi_\omega(s)$, and $\Phi_\gamma(s, \omega) = \emptyset$, otherwise.

We are now ready to define the semantics of $D$.

**Semantics of $D$.** Intuitively, we use $D$ to associate with sets of states $S$, and actions or observations $\sigma$ a probability distribution on future sets of states $Pr_\sigma(\cdot | S)$. We say $D$ is *consistent* iff $\Phi_\gamma(s, \alpha) \neq \emptyset$ for all states $s$, actions $\alpha$ with $\pi_\alpha(s)$, and contexts $\gamma \in I(\mathcal{X}_\alpha)$. We now first extend $\pi_\sigma(s)$ and $\Phi_\gamma(s, \sigma)$ from states $s$ to sets of states $S$.

In order to specify how preconditions of actions (resp., observed formulas) are evaluated on sets of states, we add modality labels to actions (resp., observations). Formally, a *labeled action* (resp., *labeled observation*) is of the form $\circ \tau$, where $\circ \in \{\diamond, \square\}$ and $\tau$ is an action (resp., observation). For sets of states $S$, we use $\pi_{\diamond \tau}(S)$ (resp., $\pi_{\square \tau}(S)$) to denote $\exists s \in S: \pi_\tau(s)$ (resp., $\forall s \in S: \pi_\tau(s)$). Moreover, we define $\mathcal{X}_{\circ \tau} = \mathcal{X}_\tau$ and $Pr_{\circ \tau} = Pr_\tau$.

For every set of states $S$, every labeled action or observation $\sigma = \circ \tau$ with $\pi_\sigma(S)$, and every context $\gamma \in I(\mathcal{X}_\sigma)$, we then define $\Phi_\gamma(S, \sigma) = \bigcup_{s \in S} \Phi_\gamma(s, \tau)$. Observe that for observations $\omega$, it holds that $\Phi_\gamma(S, \diamond \omega) = \{s \in S \mid s \models \omega\}$ and $\Phi_\gamma(S, \square \omega) = S$.

We are now ready to define the probabilistic transition between sets of states $S$ and $S'$ under $\sigma$ with $\pi_\sigma(S)$ by:

$$Pr_\sigma(S'|S) = \sum_{\gamma \in I(\mathcal{X}_\sigma), S' = \Phi_\gamma(S, \sigma)} Pr_\sigma(\gamma).$$

Intuitively, given any set of states $S$ such that $\pi_\sigma(S)$, the $\Phi_\gamma(S, \sigma)$'s are the future sets of states under $\sigma$, where each $\Phi_\gamma(S, \sigma)$ has the probability $Pr_\sigma(\gamma)$.

**Assumption 2.7** In the rest of this paper, we implicitly assume that $D$ and $D_0$ are consistent, and that all static causal laws (1) in $D$ over rigid variables also belong to $D_0$.

## 3 HISTORIES AND BELIEF STATES

Our framework for reasoning and planning in PC+ involves finite sequences of labeled actions and observations, called *histories*, which are inductively defined as follows. The *empty history* $\varepsilon$ is a history. If $h$ is a history, and $\sigma$ is a labeled action or observation, then $h, \sigma$ is a history. Histories $\varepsilon, r$ are abbreviated by $r$. The *action length* of a history $h$ is the number of occurrences of actions in $h$.

**Example 3.1** In the running example, $\Diamond goto(b), \Box pickup,$ $\Box goto(c), \Box\, at(o_1)=c \lor at(o_2)=c$ is a history of action length 3. Informally, it represents the statement "if $goto(b)$ has been executed, then $pickup, goto(c)$ can be executed, and $at(o_1)=c \lor at(o_2)=c$ is observed after that".

We use the notion of a belief state to describe the probabilistic information associated with a history $h$. Intuitively, a belief state consists of a probability value for $h$ and a probability function on a set of state sets.

**Definition 3.2** The *belief state* $b_h = (p_h, \mathcal{S}_h, Pr_h)$ for a history $h$ consists of a real number $p_h \in [0, 1]$ (called *probability* of $h$, denoted $Pr(h)$), a set of state sets $\mathcal{S}_h$, and a probability function $Pr_h$ on $\mathcal{S}_h$. It is inductively defined by:

- If $h = \varepsilon$, then $p_h = 1$, $\mathcal{S}_h = \{\Phi_\gamma \mid \gamma \in I(\mathcal{X}_0)\}$, and for all $S \in \mathcal{S}_h$, $Pr_h(S) = \sum_{\gamma \in I(\mathcal{X}_0), S=\Phi_\gamma} Pro(\gamma)$.

- If $h = r, \sigma$, then $b_h$ is defined iff (i) $b_r = (p_r, \mathcal{S}_r, Pr_r)$ is defined, and (ii) $\pi_\sigma(S)$ for some $S \in \mathcal{S}_r$. If $b_h$ is defined, then it is given as follows:

$$p_h = p_r \cdot \sum_{S \in \mathcal{S}_r, \pi_\sigma(S)} Pr_r(S)$$
$$\mathcal{S}_h = \{\Phi_\gamma(S, \sigma) \mid \gamma \in I(\mathcal{X}_\sigma), S \in \mathcal{S}_r, \pi_\sigma(S)\}$$
$$Pr_h(S') = \tfrac{p_r}{p_h} \cdot \sum_{S \in \mathcal{S}_r, \pi_\sigma(S)} Pr_\sigma(S'|S) \cdot Pr_r(S) \quad (\forall S' \in \mathcal{S}_h).$$

Intuitively, $b_\varepsilon$ describes the probabilistic knowledge associated with $D_0$, while $b_{r,\sigma}$ represents the probabilistic knowledge about the world after the history $r, \circ \alpha$ (resp., $r, \circ \omega$), which depends on (i) our respective knowledge after $r$, and (ii) the effects of $\alpha$ encoded in $D$ (resp., the observation $\omega$).

The following result shows that $\sigma = \circ\,\omega$ corresponds to a conditioning of $Pr_r$ on all state sets $S \in \mathcal{S}_r$ with $\pi_\sigma(S)$, along with removing from such $S$ all states violating $\omega$.

**Proposition 3.3** Let $h = r, \sigma$ be a history, where $\sigma = \circ\,\omega$ is a labeled observation. Then, $b_h = (p_h, \mathcal{S}_h, Pr_h)$ is defined iff (i) $b_r = (p_r, \mathcal{S}_r, Pr_r)$ is defined, and (ii) some $S \in \mathcal{S}_r$ exists with $\pi_\sigma(S)$. If $b_h$ is defined, then it is given by:

$$p_h = p_r \cdot \sum_{S \in \mathcal{S}_r, \pi_\sigma(S)} Pr_r(S)$$
$$\mathcal{S}_h = \{\{s \in S \mid s \models \omega\} \mid S \in \mathcal{S}_r, \pi_\sigma(S)\}$$
$$Pr_h(S') = \tfrac{p_r}{p_h} \cdot \sum_{\substack{S \in \mathcal{S}_r, \pi_\sigma(S), \\ S' = \{s \in S \mid s \models \omega\}}} Pr_r(S) \quad (\forall S' \in \mathcal{S}_h).$$

## 4 REASONING AND PLANNING IN $\mathcal{PC}+$

We now show at the examples of prediction, postdiction, and planning [18, 12, 7] how important problems in probabilistic reasoning about actions can be formulated in our framework of $\mathcal{PC}+$. We define them in terms of probabilities of histories in $\mathcal{PC}+$. Recall that the probability of a history $h$, denoted $Pr(h)$, is defined as $p_h$, where $b_h = (p_h, \mathcal{S}_h, Pr_h)$ is the belief state for $h$.

### 4.1 PREDICTION

We consider the following *probabilistic prediction* (or also *probabilistic temporal projection*) problem: Compute the probability that a sequence $\sigma$ of actions and observations is *certainly* possible, given that another sequence $\sigma'$ of actions and observations has occurred. Here, the probability that $\sigma$ is certainly possible after $\sigma'$ is the tight lower bound for the probability that $\sigma$ is possible after $\sigma'$. We now define this probability in terms of probabilities of histories. Note that further semantically meaningful probabilities can be defined in a similar way, for example, the probability that $\sigma$ *may be* possible after $\sigma'$, which is the tight upper bound for the probability that $\sigma$ is possible after $\sigma'$.

**Definition 4.1** Let $\sigma = \sigma_1, \ldots, \sigma_m$ and $\sigma' = \sigma'_1, \ldots, \sigma'_n$ be sequences of actions and observations. The *probability that $\sigma$ is certainly possible*, denoted $Pred(\varepsilon \triangleright \sigma)$, is defined as $Pr(\Box\sigma_1, \ldots, \Box\sigma_m)$. The *probability that $\sigma$ is certainly possible after $\sigma'$*, denoted $Pred(\sigma' \triangleright \sigma)$, is defined as $Pr(\Diamond\sigma'_1, \ldots, \Diamond\sigma'_n, \Box\sigma_1, \ldots, \Box\sigma_m) / Pr(\Diamond\sigma'_1, \ldots, \Diamond\sigma'_n)$.

**Proposition 4.2** Let $\alpha = \alpha_1, \alpha_2, \ldots, \alpha_n$ (resp., $\omega_1, \omega_2, \ldots, \omega_n$) be a sequence of actions (resp., observations), and let $\phi$ and $\psi$ be observations. Then,

- $Pred(\varepsilon \triangleright \phi, \alpha, \psi)$ is the probability that $\phi$ certainly holds initially, that $\alpha$ can certainly be executed, and that $\psi$ certainly holds after that.

- $Pred(\phi \triangleright \alpha, \psi)$ is the probability that $\alpha$ can certainly be executed, and that then $\psi$ certainly holds, given that $\phi$ is observed initially.

- $Pred(\phi, \alpha \triangleright \psi)$ is the probability that $\psi$ certainly holds, given $\phi$ is observed initially, and $\alpha$ has been executed.

- $Pred(\varepsilon \triangleright \phi, \alpha_1, \omega_1, \alpha_2, \omega_2, \ldots, \alpha_n, \omega_n)$ is the probability that $\phi$ certainly holds initially, that $\alpha = \alpha_1, \alpha_2, \ldots, \alpha_n$ can certainly be executed, and that then $\omega_1, \omega_2, \ldots, \omega_n$, respectively, certainly hold.

**Example 4.3** Let $D_0$ and $D$ be given as in Example 2.6. Suppose that the robot $r$ is initially at location $a$ and holds no object. Moreover, assume that the two objects $o_1$ and $o_2$ are both at location $b$. Then, the probability that the robot $r$ can certainly move to $b$, can certainly pickup an object, and can certainly move to $c$, and that then at least one object is certainly at location $c$ is given by 0.855: Let $\phi = at(r) = a \land at(o_1) = b \land at(o_2) = b \land holds = nil$, $\alpha = goto(b), pickup, goto(c)$, and $\psi = at(o_1) = c \lor at(o_2) = c$. We then have $Pred(\phi \triangleright \alpha, \psi) = 0.855$, which is obtained as follows (cf. Fig. 2.a). Observation $\phi$ leads to a context $(0 : c_{at(o_1)}, 0 : c_{at(o_2)}) = (b, b)$ of probability $0.8 \cdot 0.6$. Action $goto(b)$ then yields two contexts, $1 : c_g(b) = ok$ and $1 : c_g(b) = \mathit{fail}$ of probabilities 0.95 and 0.05, respectively. Action $pickup$ is only executable in $1 : c_g(b) = ok$, which leads to the empty context $\emptyset$ of probability 1. Finally, action $goto(c)$ yields two contexts $3 : c_g(c) = ok$ and $3 : c_g(c) =$

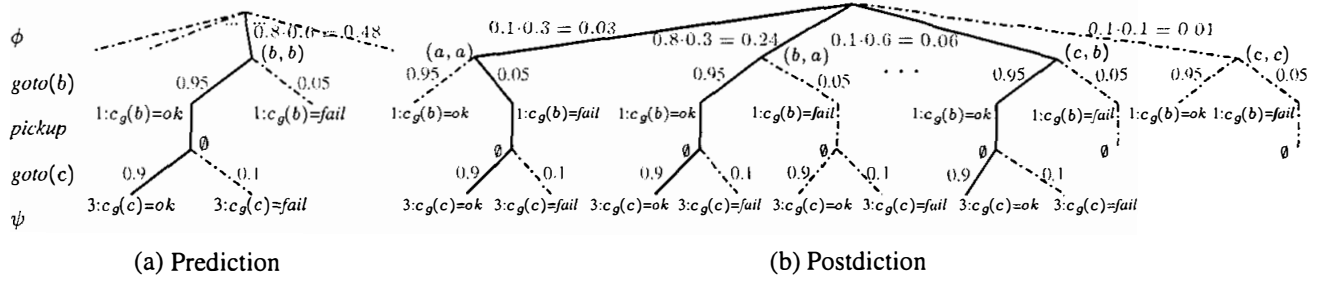

(a) Prediction                                     (b) Postdiction

Figure 2: Evolving Contexts[1]

*fail* of probabilities 0.9 and 0.1, respectively; in the former, $\psi$ is true in all states, but not in the latter. In summary, $Pred(\phi \triangleright \alpha, \psi) = Pr(\Diamond \phi, \Box \alpha, \Box \psi) / Pr(\Diamond \phi) = (0.8 \cdot 0.6 \cdot 0.95 \cdot 0.9) / (0.8 \cdot 0.6) = 0.855$.

The probability that at least one object is certainly at location $c$ after $r$ moved to $b$, picked up an object, and then moved to $c$ is given by $Pred(\phi, \alpha \triangleright \psi) = 0.9$.

The probability that $r$ can certainly move to $b$, can certainly pickup an object, and can certainly move to $c$, and that then object $o_1$ is certainly at location $c$ is given by $Pred(\phi \triangleright \alpha, \psi') = 0$, where $\psi' = at(o_1) = c$, as in no emerging context, $\psi'$ is true in all states. Indeed, in the worst case, $r$ might pick up $o_2$ and carry it to $c$. Thus, the probability that $o_1$ is at $c$ after $\phi, \alpha$ may be 0. In contrast, if we have a probability distribution for *pickup*, then $Pred(\phi \triangleright \alpha, \psi') > 0$, if $r$ picks up $o_1$ with a positive probability.

Finally, suppose that object $o_2$ is initially at either $a$ or $b$, rather than at $b$. Hence, consider now $\phi' = at(r) = a \wedge at(o_1) = b \wedge (at(o_2) = a \vee at(o_2) = b) \wedge holds = nil$, which is weaker than $\phi$. We then obtain $Pred(\phi' \triangleright \alpha, \psi) = Pred(\phi \triangleright \alpha, \psi)$ and $Pred(\phi' \triangleright \alpha, \psi') > Pred(\phi \triangleright \alpha, \psi')$. Here, we have a positive probability, since there exists a combined context of positive probability that satisfies $\phi'$, where object $o_2$ is absent from $b$, which makes *pickup* deterministic, and thus $o_1$ is certainly carried to $c$.

### 4.2 POSTDICTION

Informally, the *probabilistic postdiction* (or also *probabilistic explanation*) problem that we consider here can be formulated as follows: Compute the probability that observations were certainly holding along a sequence of actions and observations $\nu$ that actually happened.

**Definition 4.4** Let $\nu_2$ be a sequence of actions and observations, and let $\nu_1$ result from $\nu_2$ by removing some observations. Then, the probability that $\nu_2$ certainly occurred if $\nu_1$ has occurred, denoted $Post(\nu_2|\nu_1)$, is defined as $Pr(\nu_2') / Pr(\nu_1')$, where $\nu_2'$ and $\nu_1'$ result from $\nu_2$ and $\nu_1$, respectively, by labeling removed observations in $\nu_2$ with $\Box$ and all other actions and observations with $\Diamond$.

**Proposition 4.5** Let $\alpha = \alpha_1, \alpha_2, \ldots, \alpha_n$ (resp., $\omega_1, \omega_2, \ldots, \omega_n$) be a sequence of actions (resp., observations), $\nu = \alpha_1, \omega_1, \alpha_2, \omega_2, \ldots, \alpha_n, \omega_n$, and $\phi, \psi$ be observations. Then,

- $Post(\phi, \alpha, \psi | \alpha, \psi)$ is the probability that $\phi$ certainly held initially, given $\alpha$ was executed and $\psi$ was observed.

- $Post(\phi, \nu | \nu)$ is the probability that $\phi$ was certainly holding initially, given $\alpha_1, \alpha_2, \ldots, \alpha_n$ was executed, and $\omega_1, \omega_2, \ldots, \omega_n$, respectively, was observed after that.

- $Post(\phi, \nu | \alpha)$ is the probability that $\phi$ was certainly holding initially, and $\omega_1, \omega_2, \ldots, \omega_n$ was certainly holding after $\alpha_1, \alpha_2, \ldots, \alpha_n$, respectively, given $\alpha$ was executed.

**Example 4.6** Let $D_0$ and $D$ be given as in Example 2.6. Suppose that the robot $r$ moved to $b$, picked up an object, and then moved to $c$, and that the object $o_1$ was observed at $c$ after that. Then, the probability that the object $o_1$ was certainly at $b$ in the initial situation is given by 0.923: Consider $\alpha = goto(b), pickup, goto(c), \psi = at(o_1) = c$, and $\phi = at(o_1) = b$. We then have $Post(\phi, \alpha, \psi | \alpha, \psi) = 0.923$, which is obtained as follows: Each of the 9 initial contexts, given by the value pairs for $(0:c_{at(o_1)}, 0:c_{at(o_2)})$, except $(c, c)$ admits execution of $\alpha$ resulting in a context in which $\psi$ is observable. In detail, $(b, a), (b, b), (b, c),$ and $(c, b)$ can be extended by $1:c_g(b) = ok, \emptyset$, and $3:c_g(c) = ok$, and $(a, a), (a, b), (a, c),$ and $(c, a)$ by $1:c_g(b) = fail, \emptyset$, $3:c_g(c) = ok$ (cf. Fig. 2.b). Let $\nu_1 = \alpha, \psi$ and $\nu_2 = \phi, \alpha, \psi$. Then, $Pr(\nu_1') = (0.24 + 0.48 + 0.08 + 0.06) \cdot 0.95 \cdot 0.9 + (0.03 + 0.06 + 0.01 + 0.03) \cdot 0.05 \cdot 0.9 = 0.741$ and $Pr(\nu_2') = (0.24 + 0.48 + 0.08) \cdot 0.95 \cdot 0.9 = 0.684$, as $\phi$ only holds in $(b, a), (b, b),$ and $(b, c)$. Hence, $Post(\phi, \alpha, \psi | \alpha, \psi) = Pr(\nu_2') / Pr(\nu_1') = 0.923$. Note that if $at(r) \neq b$ would be observed after $goto(b)$ in $\alpha$, then we could conclude that initially $\phi$ has the probability 0, which is intuitive.

### 4.3 PLANNING

We now formulate the notions of a (sequential) plan and of its goodness for reaching a goal observation given that a sequence of actions and observations has occurred.

**Definition 4.7** Let $\nu$ be a sequence of actions and observations and $\psi$ an observation. The sequence of actions $\alpha = \alpha_1, \ldots, \alpha_n$ is a plan of goodness $g$ for $\psi$ after $\nu$ has occurred, denoted $Plan(\nu; \alpha; \psi) = g$, iff $Pred(\nu \triangleright \alpha, \psi) = g$.

---

[1]Pairs $(x, y)$ are short for $(0:c_{at(o_1)}, 0:c_{at(o_2)}) = (x, y)$.

In the general planning problem in our framework, we are then given a sequence of actions and observations $\nu$ that has occurred, a goal observation $\psi$, and a threshold $\theta$, and we want to compute a plan $\alpha$ such that $Plan(\nu; \alpha; \psi) \geqslant \theta$.

**Example 4.8** Let $D_0$ and $D$ be as in Example 2.6. Let $\phi = at(r){=}a \wedge at(o_1){=}b \wedge at(o_2){=}b$ and $\psi = at(o_1){=}c \vee at(o_2){=}c$. Then, $\alpha = goto(b), pickup, goto(c)$ is a plan of goodness 0.885 for $\psi$ given that $\phi$ holds initially.

On the other hand, $Plan(\phi; \alpha; \psi') = 0$ for $\psi' = at(o_1){=}c$, since $r$ might (unpredictably) carry $o_2$ to $c$ instead of $o_1$. However, $Plan(\phi, \alpha', \psi')=0.885^2=0.731$ for $\alpha'=\alpha, drop, goto(b), pickup, goto(c)$. Note that $\alpha'$ is optimal, since moving twice to $c$ and to $b$ is necessary in general.

If $pickup$ would be probabilistic and, e.g., obey the uniform distribution, then $Plan(\phi; \alpha; \psi') > 0$ would hold. Indeed, there would be a context $c$ after executing $\alpha$ where $\psi$ holds in all its associated states. A wrong $pickup$ decreases the success probability, which is, however, still non-zero.

## 5 SUMMARY AND OUTLOOK

We have presented the probabilistic action language $PC+$, which generalizes $C+$ by probabilistic information into an expressive framework for dealing with qualitative as well as quantitative uncertainty about actions. Its formal semantics is defined in terms of probabilistic transitions between sets of states. We have then shown that, using the concepts of a history and its belief state, several important problems in reasoning about actions can be concisely expressed in it.

In the extended report [4], we also provide a formulation of conditional planning in $PC+$. Furthermore, we present a compact representation of belief states, which is based on the notion of a context to encode possible sets of states, and we prove its correctness in implementing belief states. Finally, we also discuss how to reduce probabilistic reasoning about actions in $PC+$ to reasoning in nonmonotonic causal theories [7], which is a step towards implementation on top of existing technology for nonmonotonic causal theories (such as the Causal Calculator [14]).

An interesting topic of future research is to provide more efficient algorithms and a detailed complexity analysis for probabilistic reasoning about actions in $PC+$. Other interesting topics are to add costs to planning and conditional planning and to define in our framework further semantic notions like counterfactuals, interventions, actual cause, and causal explanations, taking inspiration by similar concepts in the structural-model approach to causality [15, 9].

### Acknowledgments

This work was partially supported by FWF (Austrian Science Funds) under the projects P14781-INF, P16536-N04, and Z29-N04 and by the European Commission under the grant FET-2001-37004 WASP and the Marie Curie Individual Fellowship HPMF-CT-2001-001286 of the programme "Human Potential" (Disclaimer: The authors are solely responsible for information communicated and the European Commission is not responsible for any views or results expressed). We are grateful to the reviewers for their constructive comments, which helped to improve our work.